\pdfoutput=1
\documentclass[11pt,a4paper]{article}
\usepackage[table]{xcolor}
\usepackage[hyperref]{acl2020}
\usepackage[T1]{fontenc}
\usepackage{times}
\usepackage{latexsym}

\usepackage{etoolbox}
\usepackage{pbox}
\usepackage{array}
\usepackage{amsmath}
\usepackage{amssymb}
\usepackage{booktabs}
\usepackage{relsize}
\usepackage{enumitem}
\setlist[itemize]{label=$\triangleright$,nolistsep,itemsep=0.2em,leftmargin=1.5em}

\usepackage{tcolorbox}
\tcbuselibrary{breakable}
{\begin{tcolorbox}[breakable,parbox=false,notitle,boxrule=0.5pt,colback=orange!10,colframe=orange,arc=0pt,boxsep=0mm,left=2pt,right=2pt,top=2pt,bottom=2pt]
\textbf{C.S.}
}{\end{tcolorbox}}

\usepackage{microtype}
\aclfinalcopy

\makeatletter

\DeclareRobustCommand\citepos
{\begingroup
	\let\NAT@nmfmt\NAT@posfmt
	\NAT@swafalse\let\NAT@ctype\z@\NAT@partrue
	\@ifstar{\NAT@fulltrue\NAT@citetp}{\NAT@fullfalse\NAT@citetp}}

\let\NAT@orig@nmfmt\NAT@nmfmt
\def\NAT@posfmt#1{\NAT@orig@nmfmt{#1's}}

\makeatother
\usepackage{quoting}

\newcommand{\citequote}[2]{\begin{quoting}[indentfirst=false,leftmargin=0.5cm,rightmargin=0.5cm,font={raggedright}] \smaller\emph{``#1''}\hfill\mbox{\textrm{---\citet{#2}}}\end{quoting}\vspace{-1em}\noindent}

\newcommand{\altcitequote}[2]{\begin{quoting}[indentfirst=false,leftmargin=0.5cm,rightmargin=0.5cm,font={raggedright}] \smaller\emph{``#1''\\}\hfill\mbox{\textrm{---\citet{#2}}}\end{quoting}\vspace{-1em}\noindent}

\newcommand{\citequotenoattr}[2]{\begin{quoting}[indentfirst=false,leftmargin=0.5cm,rightmargin=0.5cm,font={raggedright}] \smaller\emph{``#1''}\end{quoting}\vspace{-1em}\noindent}

\title{Language (Technology) is Power: A Critical Survey of ``Bias'' in NLP}

\author{Su Lin Blodgett \\
	College of Information and Computer Sciences \\
	University of Massachusetts Amherst \\
	\texttt{blodgett@cs.umass.edu} \\\And
	Solon Barocas \\
	Microsoft Research \\
	Cornell University \\
	\texttt{solon@microsoft.com} \\\AND
	Hal Daum\'{e} III \\
	Microsoft Research \\
	University of Maryland \\
	\texttt{me@hal3.name} \\\And
	Hanna Wallach \\
	Microsoft Research \\
	\texttt{wallach@microsoft.com} \\}

\begin{document}

\maketitle
\begin{abstract}
  We survey 146 papers analyzing ``bias'' in NLP systems, finding that their motivations are often vague, inconsistent, and lacking in normative reasoning, despite the fact that analyzing ``bias'' is an inherently normative process. We further find that these papers' proposed quantitative techniques for measuring or mitigating ``bias'' are poorly matched to their motivations and do not engage with the relevant literature outside of NLP. Based on these findings, we describe the beginnings of a path forward by proposing three recommendations that should guide work analyzing ``bias'' in NLP systems. These recommendations rest on a greater recognition of the relationships between language and social hierarchies, encouraging researchers and practitioners to articulate their conceptualizations of ``bias''---i.e., what kinds of system behaviors are harmful, in what ways, to whom, and why, as well as the normative reasoning underlying these statements---and to center work around the lived experiences of members of communities affected by NLP systems, while interrogating and reimagining the power relations between~technologists and such communities.\looseness=-1
\end{abstract}

\section{Introduction}
\label{s:intro}
A large body of work analyzing ``bias'' in natural language processing (NLP) systems has emerged in recent years, including work on ``bias'' in embedding spaces \cite[e.g.,][]{bolukbasi2016man,caliskan2017semantics,gonen2019lipstick,may2019measuring} as well as work on ``bias'' in systems developed for a breadth of tasks including language modeling \cite{lu2018gender,bordia2019identifying}, coreference resolution \cite{rudinger2018gender,zhao2018gender}, machine translation \cite{vanmassenhove2018getting,stanovsky2019evaluating}, sentiment analysis \cite{kiritchenko2018examining}, and hate speech/toxicity detection \cite[e.g.,][]{park2018reducing,dixon2018measuring}, among others.\looseness=-1

Although these papers have laid vital groundwork by illustrating some of the ways that NLP systems can be harmful, the majority of them fail to engage critically with what constitutes ``bias'' in the first place. Despite the fact that analyzing ``bias'' is an inherently normative process---in which some system behaviors are deemed good and others harmful---papers on ``bias'' in NLP systems are rife with unstated assumptions about what kinds of system behaviors are harmful, in what ways, to whom, and why. Indeed, the term ``bias'' (or ``gender bias'' or ``racial bias'') is used to describe a wide range of system behaviors, even though they may be harmful in different ways, to different groups, or for different reasons. Even papers analyzing ``bias'' in NLP systems developed for the same task often conceptualize it differently.\looseness=-1

For example, the following system behaviors are all understood to be self-evident statements of ``racial bias'': (a)
embedding spaces in which embeddings for names associated with African Americans are closer (compared to names associated with European Americans) to unpleasant words than pleasant words  \cite{caliskan2017semantics}; (b) sentiment analysis systems yielding different intensity scores for sentences containing names associated with African Americans and sentences containing names associated with European Americans \cite{kiritchenko2018examining}; and (c) toxicity detection systems scoring tweets containing features associated with African-American English as more offensive than tweets without these features \cite{davidson2019racial,sap2019risk}. Moreover, some of these papers focus on ``racial bias'' expressed in written text, while others focus on ``racial bias'' against authors. This use of imprecise terminology obscures these important differences.\looseness=-1

We survey 146 papers analyzing ``bias'' in NLP systems, finding that their motivations are often vague and inconsistent. Many lack any normative reasoning for why the system behaviors that are described as ``bias'' are harmful, in what ways, and to whom. Moreover, the vast majority of these papers do not engage with the relevant literature outside of NLP to ground normative concerns when proposing quantitative techniques for measuring or mitigating ``bias.'' As a result, we find that many of these techniques are poorly matched to their motivations, and are not comparable to one another.\looseness=-1

We then describe the beginnings of a path forward by proposing three recommendations that should guide work analyzing ``bias'' in NLP systems. We argue that such work should examine the relationships between language and social hierarchies; we call on researchers and practitioners conducting such work to articulate their conceptualizations of ``bias'' in order to enable conversations about what kinds of system behaviors are harmful, in what ways, to whom, and why; and we recommend deeper engagements between technologists and communities affected by NLP systems. We also provide several concrete research questions that are implied by each of our recommendations.\looseness=-1

\section{Method} \label{s:papers}

Our survey includes all papers known to us analyzing ``bias'' in NLP systems---146 papers in total. We omitted papers about speech, restricting our survey to papers about written text only. To identify the 146 papers, we first searched the ACL Anthology\footnote{\url{https://www.aclweb.org/anthology/}} for all papers with the keywords ``bias'' or ``fairness''  that were made available prior to May 2020. We retained all papers about social ``bias,'' and discarded all papers about other definitions of the keywords (e.g., hypothesis-only bias, inductive bias,  media bias). We also discarded all papers using ``bias'' in NLP systems to measure social ``bias'' in text or the real world \cite[e.g.,][]{garg2018word}.\looseness=-1

To ensure that we did not exclude any relevant papers without the keywords ``bias'' or ``fairness,'' we also traversed the citation graph of our initial set of papers, retaining any papers analyzing ``bias'' in NLP systems that are cited by or cite the papers in our initial set. Finally, we manually inspected any papers analyzing ``bias''
in NLP systems from leading machine learning, human--computer interaction, and web conferences and workshops, such as ICML, NeurIPS, AIES, FAccT, CHI, and WWW, along with any relevant papers that were made available in the ``Computation and Language'' and ``Computers and Society'' categories on arXiv prior to May 2020, but found that they had already been identified via our traversal of the citation graph. We provide a list of all 146 papers in the appendix. In Table~\ref{t:subfields}, we provide a breakdown of the NLP tasks covered by the papers. We note that counts do not sum to 146, because some papers cover multiple tasks. For example, a paper might test the efficacy of a technique for mitigating ``bias'' in embedding~spaces in the context of sentiment analysis.\looseness=-1

\begin{table}[t]
	\centering
		\begin{footnotesize}
			\begin{tabular}{rc}
			    \toprule
				\textbf{NLP task} & \textbf{Papers}  \\
				\midrule
				Embeddings (type-level or contextualized) & 54 \\
				Coreference resolution & 20 \\
				Language modeling or dialogue generation & 17 \\
				Hate-speech detection & 17 \\
				Sentiment analysis & 15 \\
				Machine translation & \hphantom{0}8 \\
				Tagging or parsing & \hphantom{0}5 \\
				Surveys, frameworks, and meta-analyses & 20  \\
				Other & 22  \\
				\bottomrule
			\end{tabular}
			\caption{The NLP tasks covered by the 146 papers.\looseness=-1}
			\label{t:subfields}
		\end{footnotesize}
\end{table}

Once identified, we then read each of the 146 papers with the goal of categorizing their motivations and their proposed quantitative techniques for measuring or mitigating ``bias.'' We used a previously developed taxonomy of harms for this categorization, which differentiates between so-called \emph{allocational} and \emph{representational} harms \cite{barocas2017problem,crawford2017trouble}. Allocational harms arise when an automated system allocates resources (e.g., credit) or opportunities (e.g., jobs) unfairly to different social groups; representational harms arise when a system (e.g., a search engine) represents some social groups in a less favorable light than others, demeans them, or fails to recognize their existence altogether. Adapting and extending this taxonomy, we categorized the 146 papers' motivations and techniques into the following categories:\looseness=-1
\begin{itemize}
	\item \emph{Allocational harms.}
	\item \emph{Representational harms:}\footnote{We grouped several types of representational harms into two categories to reflect that the main point of differentiation between the 146 papers' motivations and proposed quantitative techniques for measuring or mitigating ``bias'' is whether or not they focus on stereotyping. Among the papers that do not focus on stereotyping, we found that most lack sufficiently clear motivations and techniques to reliably categorize them further.\looseness=-1}
	    \begin{itemize}
	    \item \emph{Stereotyping} that propagates negative generalizations about particular social groups.
	    \item Differences in \emph{system performance} for different social groups, language that \emph{misrepresents} the distribution of different social groups in the population, or language that is \emph{denigrating} to particular social groups.
	    \end{itemize}
	\item \emph{Questionable correlations} between system behavior and features of language that are typically associated with particular social groups.
	\item \emph{Vague descriptions} of ``bias'' (or ``gender bias''~or ``racial bias'') or \emph{no description} at all.
	\item \emph{Surveys, frameworks, and meta-analyses}.
\end{itemize}

\begin{table}[t]
	\begin{center}
		\begin{footnotesize}
			\begin{tabular}{>{\raggedleft}p{0.23\textwidth}cc}
			    \toprule
			    & \multicolumn{2}{c}{\textbf{Papers}} \\
			    \cmidrule{2-3}
				\textbf{Category} & \textbf{Motivation} & \textbf{Technique} \\
				\midrule
				Allocational harms & 30 & \hphantom{0}4 \\
				Stereotyping & 50 & 58 \\
				Other representational harms & 52 & 43 \\
			    Questionable correlations & 47 & 42 \\
				Vague/unstated & 23 & \hphantom{0}0 \\
				Surveys, frameworks, and meta-analyses & 20 & 20 \\
				\bottomrule
			\end{tabular}
			\caption{The categories into which the 146 papers fall.}
			\label{t:stats}
		\end{footnotesize}
	\end{center}
\end{table}

In Table \ref{t:stats} we provide counts for each of the six categories listed above. (We also provide a list of the papers that fall into each category in the appendix.) Again, we note that the counts do not sum to 146, because some papers state multiple motivations, propose multiple techniques, or propose a single technique for measuring or mitigating multiple harms. Table \ref{t:quotes}, which is in the appendix, contains examples of the papers' motivations and techniques across a range of different NLP tasks.\looseness=-1

\section{Findings}
\label{s:findings}

Categorizing the 146 papers' motivations and proposed
quantitative techniques for measuring or mitigating ``bias'' into the
six categories listed above enabled us to identify several
commonalities, which we present below, along with illustrative
quotes.\looseness=-1

\subsection{Motivations}

\paragraph{Papers state a wide range of motivations, multiple motivations, vague motivations, and sometimes no motivations at all.}
We found that the papers' motivations span all six categories, with several papers falling into each one. Appropriately, papers that provide surveys or frameworks for analyzing ``bias'' in NLP systems often state multiple motivations \citep[e.g.,][]{hovy2016social,bender2019typology,sun2019mitigating,rozado2020wide,shah2020predictive}. However, as the examples in Table \ref{t:quotes} (in the appendix) illustrate, many other papers (33\%) do so as well.
Some papers (16\%) state only vague motivations or no motivations at all. For example,\looseness=-1

\altcitequote{[N]o human should be discriminated on the basis of demographic attributes by an NLP system.}{kaneko2019gender}
\vspace{-.5em}
\citequote{[P]rominent word embeddings [...] encode systematic biases against women and black people [...] implicating many NLP systems in scaling up social injustice.}{may2019measuring}

\noindent These examples leave unstated what it might mean for an NLP system to ``discriminate,'' what constitutes ``systematic biases,'' or how NLP systems contribute to ``social injustice'' (itself undefined).\looseness=-1

\paragraph{Papers' motivations sometimes include no normative reasoning.} We found that some papers (32\%) are not motivated by any apparent normative concerns, often focusing instead on concerns about system performance. For example, the first quote below includes normative reasoning---namely that models should not use demographic information to make predictions---while the other focuses on learned correlations impairing system performance.\looseness=-1

\citequote{In [text classification], models are expected to make predictions with the semantic information rather than with the demographic group identity information (\emph{e.g.,} `gay', `black') contained in the sentences.}{zhang2020demographics}

\altcitequote{An over-prevalence of some gendered forms in the training data leads to translations with identifiable errors. Translations are better for sentences involving men and for sentences containing stereotypical gender roles.}{saunders2020reducing}

\paragraph{Even when papers do state clear motivations, they are often unclear about why the system behaviors that are described as ``bias'' are harmful, in what ways, and to whom.} We found that even papers with clear motivations often fail to explain what kinds of system behaviors are harmful, in what ways, to whom, and why.
For example,\looseness=-1

\citequote{Deploying these word embedding algorithms in practice, for example in automated translation systems or as hiring aids, runs the serious risk of perpetuating problematic biases in important societal contexts.}{brunet2019understanding}

\citequote{[I]f the systems show discriminatory behaviors in the interactions, the user experience will be adversely affected.}{liu2019does}

\noindent These examples leave unstated what ``problematic biases'' or non-ideal user experiences might look like, how the system behaviors might result in these things, and who the relevant stakeholders or users might be. In
contrast, we find that papers that provide surveys or frameworks for analyzing
``bias'' in NLP systems often name who is harmed, acknowledging that
different social groups may experience these systems differently due
to their different relationships with NLP systems or different social
positions. For example, \citet{ruane2019conversational} argue for a
``deep understanding of the user groups [sic] characteristics,
contexts, and interests'' when designing conversational
agents.\looseness=-1

\paragraph{Papers about NLP systems developed for the same task often conceptualize ``bias'' differently.} Even papers that cover the same NLP task often conceptualize ``bias'' in ways that differ substantially and are sometimes inconsistent. Rows 3 and 4 of Table \ref{t:quotes} (in the appendix) contain machine translation papers with different conceptualizations of ``bias,'' leading to different proposed techniques, while rows 5 and 6 contain papers on ``bias'' in embedding spaces that state different motivations, but propose techniques for quantifying stereotyping.\looseness=-1

\paragraph{Papers' motivations conflate allocational and representational harms.} We found that the papers' motivations sometimes (16\%) name immediate representational harms, such as stereotyping, alongside more distant allocational harms, which, in the case of stereotyping, are usually imagined as downstream effects of stereotypes on r\'{e}sum\'{e} filtering. Many of these papers use the imagined downstream effects to justify focusing on particular system behaviors, even when the downstream effects are not measured. Papers on ``bias'' in embedding spaces are especially likely to do this because embeddings are often used as input to other systems:\looseness=-1

\altcitequote{However, none of these papers [on embeddings] have recognized how blatantly sexist the embeddings are and hence risk introducing biases of various types into real-world systems.}{bolukbasi2016man}
\vspace{-.5em}
\citequote{It is essential to quantify and mitigate gender bias in these embeddings to avoid them from affecting downstream applications.}{zhou2019examining}

\noindent In contrast, papers that provide surveys or frameworks for analyzing ``bias'' in NLP systems treat representational harms as harmful in their own right. For example, \citet{mayfield2019equity} and \citet{ruane2019conversational} cite the harmful reproduction of dominant linguistic norms by NLP systems (a point to which we return in section~\ref{s:path}), while \citet{bender2019typology} outlines a range of harms, including seeing stereotypes in search results and being made invisible to search engines due to language practices.\looseness=-1

\subsection{Techniques}

\paragraph{Papers' techniques are not well grounded in the relevant literature outside of NLP.} Perhaps unsurprisingly given that the papers' motivations are often vague, inconsistent, and lacking in normative reasoning, we also found that the papers' proposed quantitative techniques for measuring or mitigating ``bias'' do not effectively engage with the relevant literature outside of NLP. Papers on stereotyping are a notable exception: the Word Embedding Association Test \cite{caliskan2017semantics} draws on the Implicit Association Test  \cite{greenwald1998measuring} from the social psychology literature, while several techniques operationalize the well-studied ``Angry Black Woman'' stereotype \cite{kiritchenko2018examining,may2019measuring,tan2019assessing} and the ``double bind'' faced by women \cite{may2019measuring,tan2019assessing}, in which women who succeed at stereotypically male tasks are perceived to be less likable than similarly successful men \cite{heilman2004penalties}. \citet{tan2019assessing} also examine the compounding effects of race and gender, drawing on Black feminist scholarship on intersectionality \cite{crenshaw1989demarginalizing}.\looseness=-1

\paragraph{Papers' techniques are poorly matched to their motivations.} We found that although 21\% of the papers include allocational harms in their motivations, only four papers actually propose techniques for measuring or mitigating allocational harms. 
\looseness=-1

\paragraph{Papers focus on a narrow range of potential sources of ``bias.''}
We found that nearly all of the papers focus on system predictions as the potential sources of ``bias,'' with many additionally focusing on ``bias'' in datasets (e.g., differences in the number of gendered pronouns in the training data \cite{zhao2019gender}). Most papers do not interrogate the normative implications of other decisions made during the development and deployment lifecycle---perhaps unsurprising given that their motivations sometimes include no normative reasoning. A few papers are exceptions, illustrating the impacts of task definitions, annotation guidelines, and evaluation metrics: \citet{cao2019toward} study how folk conceptions of gender \cite{keyes2018misgendering} are reproduced in coreference resolution systems that assume a strict gender dichotomy, thereby maintaining cisnormativity; \citet{sap2019risk} focus on the effect of priming annotators with information about possible dialectal differences when asking them to apply toxicity labels to sample tweets, finding that annotators who are primed are significantly less likely to label tweets containing features associated with African-American English as offensive.\looseness=-1

\section{A path forward} \label{s:path}

We now describe how researchers and practitioners conducting work
analyzing ``bias'' in NLP systems might avoid the pitfalls presented
in the previous section---the beginnings of a path forward. We propose
three recommendations that should guide such work, and, for each,
provide several concrete research questions. We emphasize that
these questions are not comprehensive, and are intended to generate
further questions and lines of engagement.\looseness=-1

Our three recommendations are as follows:
\begin{enumerate}[label=\bfseries (R\arabic*),leftmargin=*]
    \itemsep0em
    \item Ground work analyzing ``bias'' in NLP systems in the relevant literature outside of NLP that explores the relationships between language and social hierarchies. Treat representational harms as harmful in their own right.\looseness=-1 \label{p:patterns}
    \item Provide explicit statements of why the system behaviors that are described as ``bias'' are harmful, in what ways, and to whom. Be forthright about the normative reasoning \cite{green2019good} underlying these statements.\label{p:name}
	\item Examine language use in practice by engaging with the lived experiences of members of communities affected by NLP systems. Interrogate and reimagine the power relations between technologists and such communities. \label{p:practice}
\end{enumerate}

\subsection{Language and social hierarchies} \label{s:relationships}

Turning first to \ref{p:patterns}, we argue that work analyzing
``bias'' in NLP systems will paint a much fuller picture
if it engages with the relevant literature outside of
NLP that explores the relationships between language and social
hierarchies. Many disciplines, including sociolinguistics, linguistic
anthropology, sociology, and social psychology, study how language takes on social meaning and the role that language plays in maintaining social hierarchies. For example, language is the means through which social groups are labeled and one way that beliefs about social groups are transmitted \citep[e.g.,][]{maass1999linguistic,beukeboom2019how}. Group labels can serve as the basis of stereotypes and thus reinforce social inequalities: ``[T]he label content functions to identify a given category of people, and thereby conveys category boundaries and a position in a hierarchical taxonomy'' \cite{beukeboom2019how}. Similarly, ``controlling images,'' such as stereotypes of Black women, which are linguistically and visually transmitted through literature, news media, television, and so forth, provide ``ideological justification'' for their continued oppression \cite[Chapter~4]{collins2000black}.\looseness=-1

As a result, many groups have sought to bring about social changes through changes in language, disrupting patterns of oppression and marginalization via so-called ``gender-fair'' language \cite{sczesny2016can,menegatti17gender}, language that is more inclusive to people with disabilities \cite{adaguidelines}, and language that is less dehumanizing (e.g., abandoning the use of the term ``illegal'' in everyday discourse on immigration in the U.S. \cite{rosa2019contesting}).
The fact that group labels are so contested is evidence of how deeply intertwined language and social hierarchies are. Taking ``gender-fair'' language as an example, the hope is that reducing asymmetries in language about women and men will reduce asymmetries in their social standing.
Meanwhile, struggles over language use often arise from dominant social groups' desire to ``control both material and symbolic resources''---i.e., ``the right to decide what words will mean and to control those meanings''---as was the case in some white speakers' insistence on using offensive place names against the objections of Indigenous speakers \cite[Chapter~3]{hill2008everyday}.\looseness=-1

Sociolinguists and linguistic anthropologists have also examined language attitudes and language ideologies, or people's metalinguistic beliefs about language: Which language varieties or practices are taken as standard, ordinary, or unmarked? Which are considered correct, prestigious, or appropriate for public use, and which are considered incorrect, uneducated, or offensive \cite[e.g.,][]{campbell2009nature,preston2009are,loudermilk2015implicit,lanehart2018black}? Which are rendered invisible \cite{roche2019articulating}?\footnote{Language ideologies encompass much more than this; see, e.g., \citet{lippi2012english}, \citet{alim2016raciolinguistics}, \citet{rosa2017unsettling}, \citet{rosa2017language}, and \citet{hudley2017language}.\looseness=-1} Language ideologies play a vital role in reinforcing and justifying social hierarchies because beliefs about language varieties or practices often translate into beliefs about their speakers \cite[e.g.][]{alim2016raciolinguistics,rosa2017unsettling,craft2020language}. For example, in the U.S., the portrayal of non-white speakers' language varieties and practices as linguistically deficient helped to justify violent European colonialism, and today continues to justify enduring racial hierarchies by maintaining views of non-white speakers as lacking the language ``required for complex thinking processes and successful engagement in~the global economy'' \cite{rosa2017unsettling}.\looseness=-1

Recognizing the role that language plays in maintaining social hierarchies is critical to the future of work analyzing ``bias'' in NLP systems. First, it helps to explain why representational harms are harmful in their own right. Second, the complexity of the relationships between language and social hierarchies illustrates why studying ``bias'' in NLP systems is so challenging, suggesting that researchers and practitioners will need to move beyond existing algorithmic fairness techniques. We argue that work must be grounded in the relevant literature outside of NLP that examines the relationships between language and social hierarchies; without this grounding, researchers and practitioners risk measuring or mitigating only what is convenient to measure or mitigate, rather~than what is most normatively concerning.\looseness=-1

More specifically, we recommend that work analyzing ``bias'' in NLP systems be reoriented around the following question: How are social hierarchies, language ideologies, and NLP systems coproduced? This question mirrors \citepos{benjamin2020vision} call to examine how ``race and technology are coproduced''---i.e., how racial hierarchies, and the ideologies and discourses that maintain them, create and are re-created by technology. We recommend that researchers and practitioners similarly ask how existing social hierarchies and language ideologies drive the development and deployment of NLP systems, and how these systems therefore reproduce these hierarchies and ideologies. As a starting point for reorienting work analyzing ``bias'' in NLP systems around this question, we provide~the following concrete research questions:\looseness=-1

\begin{itemize}
    \item How do social hierarchies and language ideologies influence the decisions made during the development and deployment lifecycle? What kinds of NLP systems do these decisions result in, and what kinds do they foreclose?\looseness=-1
    \begin{itemize}[label=$\diamond$]
        \item General assumptions: To which linguistic norms do NLP systems adhere \cite{bender2019typology,ruane2019conversational}? Which language practices are implicitly assumed to be standard, ordinary, correct, or appropriate?
        \item Task definition: For which speakers are NLP systems (and NLP resources) developed? (See \citet{joshi2020state} for a discussion.)
        How do task definitions discretize the world? For example, how are social groups delineated when defining demographic attribute prediction tasks \cite[e.g.,][]{koppel2002automatically,rosenthal2011age,nguyen2013how}? What about languages in native language prediction tasks \cite{tetreault2013report}?
        \item Data: How are datasets collected, preprocessed, and labeled or annotated? What are the impacts of annotation guidelines, annotator assumptions and perceptions \cite{olteanu2019social,sap2019risk,geiger2020garbage}, and annotation aggregation processes \cite{pavlick2019inherent}?
        \item Evaluation: How are NLP systems evaluated? What are the impacts of evaluation metrics \cite{olteanu2017limits}? Are any non-quantitative evaluations performed?
    \end{itemize}
  \item How do NLP systems reproduce or transform language ideologies? Which language varieties or practices come to be deemed good or bad? Might ``good'' language simply mean language that is easily handled by existing NLP systems? For example, linguistic phenomena arising from many language practices \cite{eisenstein2013bad} are described as ``noisy text'' and often viewed as a target for ``normalization.'' How do the language ideologies that are reproduced by~NLP systems maintain social hierarchies?\looseness=-1
  \item Which representational harms are being measured or mitigated? Are these the most normatively concerning harms, or merely those that are well handled by existing algorithmic fairness techniques? Are there other representational harms that might be analyzed?
\end{itemize}

\subsection{Conceptualizations of ``bias''}
\label{s:naming}

Turning now to \ref{p:name}, we argue that work analyzing ``bias'' in NLP systems should provide explicit statements of why the system
behaviors that are described as ``bias'' are harmful, in what ways, and to whom, as well as the normative reasoning underlying these statements. In other words, researchers and practitioners should articulate their conceptualizations
of ``bias.'' As we described above, papers often contain descriptions of system behaviors that are understood to be self-evident statements of ``bias.''
This use of imprecise terminology has led to papers all claiming to
analyze ``bias'' in NLP systems, sometimes even in systems developed
for the same task, but with different or even inconsistent
conceptualizations of ``bias,'' and no explanations for these
differences.\looseness=-1

Yet analyzing ``bias'' is an inherently normative
process---in which some system behaviors are deemed good and others
harmful---even if assumptions about what kinds of system
behaviors are harmful, in what ways, for whom, and why are not stated. We therefore echo calls by \citet{bardzell2011towards}, \citet{keyes2019human}, and \citet{green2019good} for researchers and practitioners to make their normative reasoning explicit by articulating the social values that underpin their decisions to deem some system behaviors as harmful, no matter how obvious such values appear to be. We further argue that this reasoning should take into account the relationships between language and social hierarchies that we described above. First, these relationships provide a foundation from which to approach the normative reasoning that we recommend making explicit. For example, some system behaviors might be harmful precisely because they maintain social hierarchies. Second, if work analyzing ``bias'' in NLP systems is reoriented to understand how social hierarchies, language ideologies, and NLP systems are coproduced, then this work will be incomplete if we fail to account for the ways that social hierarchies and language ideologies determine what we mean by ``bias'' in the first place. As a starting point, we therefore provide the following concrete research questions:\looseness=-1

\begin{itemize}
\item What kinds of system behaviors are described as ``bias''? What are their potential sources (e.g., general assumptions, task definition, data)?
\item In what ways are these system behaviors harmful, to whom are they harmful, and why?
\item What are the social values (obvious or not) that underpin this conceptualization of ``bias?''
\end{itemize}

\subsection{Language use in practice} \label{s:practice}

Finally, we turn to \ref{p:practice}. Our perspective, which rests on a greater recognition of the relationships between language and social hierarchies, suggests several directions for examining language use in practice. Here, we focus on two. First, because language is necessarily situated, and because different social groups have different lived experiences due to their different social positions \cite{hanna2020towards}---particularly groups at the intersections of multiple axes of oppression---we recommend that researchers and practitioners center work analyzing ``bias'' in NLP systems around the lived experiences of members of communities affected by these systems. Second, we recommend that the power relations between technologists and such communities be interrogated and reimagined. Researchers have pointed out that algorithmic fairness techniques, by proposing incremental technical mitigations---e.g., collecting new datasets or training better models---maintain these power relations by (a) assuming that automated systems should continue to exist, rather than asking whether they should be built at all, and (b) keeping development and deployment decisions in the hands of technologists \cite{bennett2019what,cifor2019feminist,green2019good,katell2020toward}.\looseness=-1

There are many disciplines for researchers and practitioners to draw on when pursuing these directions. For example, in human--computer interaction, \citet{hamidi2018gender} study transgender people's experiences with automated gender recognition systems in order to uncover how these systems reproduce structures of transgender exclusion by redefining what it means to perform gender ``normally.'' Value-sensitive design provides a framework for accounting for the values of different stakeholders in the design of technology \cite[e.g.,][]{friedman2006value,friedman2019value,ledantec2009values,yoo2019service}, while participatory design seeks to involve stakeholders in the design process itself \cite{sanders2002from,muller2007participatory,simonsen2013routledge,disalvo2013communities}. Participatory action research in education \cite{kemmis2006participatory} and in language documentation and reclamation \cite{junker2008participatory} is also relevant. In particular, work on language reclamation to support decolonization and tribal sovereignty \cite{leonard2012reframing} and work in sociolinguistics focusing on developing co-equal research relationships with community members and supporting linguistic justice efforts \cite[e.g.,][]{bucholtz2014sociolinguistic,bucholtz2016beyond,bucholtz2019california} provide examples of more emancipatory relationships with communities. Finally, several workshops and events have begun to explore how to empower stakeholders in the development and deployment of technology \cite{vaccaro2019contestability,givens2020centering,sassaman2020creating}\footnote{Also \url{https://participatoryml.github.io/}} and how to help researchers and practitioners consider when not to build systems at all \cite{barocas2020when}.\looseness=-1

As a starting point for engaging with communities affected by NLP systems, we therefore provide~the following concrete research questions:\looseness=-1
\begin{itemize}
    \item How do communities become aware of NLP systems? Do they resist them, and if so, how?
	\item What additional costs are borne by communities for whom NLP systems do not work well?
	\item Do NLP systems shift power toward oppressive institutions (e.g., by enabling predictions that communities do not want made, linguistically based unfair allocation of resources or opportunities \cite{rosa2017unsettling}, surveillance, or censorship), or away from such institutions?\looseness=-1
	\item Who is involved in the development and deployment of NLP systems? How do decision-making processes maintain power relations between technologists and communities affected by NLP systems? Can these processes~be changed to reimagine these relations?\looseness=-1
\end{itemize}

\section{Case study}

To illustrate our recommendations, we present a case study covering work on African-American English (AAE).\footnote{This language variety has
  had many different names over the years, but is now generally called African-American English
  (AAE), African-American Vernacular English (AAVE),~or
  African-American Language (AAL)
  \cite{green2002african,wolfram2015american,rickford2016language}.} Work analyzing ``bias'' in the context of AAE has shown that part-of-speech taggers, language identification systems, and
dependency parsers all work less well on text containing features associated with AAE than on text without these features \cite{jorgensen2015challenges,jorgensen2016learning,blodgett2016demographic,blodgett2018twitter},
and that toxicity detection systems score tweets containing features associated with AAE as more offensive than tweets without~them
\cite{davidson2019racial,sap2019risk}.\looseness=-1

These papers have been critical for
highlighting AAE as a language variety for which existing NLP systems
may not work, illustrating their limitations. However, they do not
conceptualize ``racial bias'' in the same way. The first four of these
papers simply focus on system performance differences between text containing features associated with AAE and text without these features. In contrast, the last two papers also focus on such system performance differences, but motivate this focus with the following additional reasoning: If tweets containing features associated with AAE are scored as more offensive than tweets without these features, then this might (a) yield negative perceptions of AAE; (b) result in disproportionate removal of tweets containing these features, impeding participation in online platforms and reducing the space available online in which speakers can use AAE freely; and (c) cause AAE speakers to incur additional costs if they have to change their language practices to~avoid negative perceptions or tweet removal.\looseness=-1

More importantly, none of these papers engage with the literature on AAE, racial hierarchies in the U.S., and raciolinguistic ideologies. By failing to engage with this literature---thereby treating AAE simply as one of many non-Penn Treebank varieties of English or perhaps as another challenging domain---work analyzing ``bias'' in NLP systems in the context of AAE fails to situate these systems in the world. Who are the speakers of AAE? How are they viewed? We argue that AAE as a language variety cannot be separated from its speakers---primarily Black people in the U.S., who experience systemic anti-Black racism---and the language ideologies that reinforce and justify racial hierarchies.\looseness=-1

Even after decades of sociolinguistic efforts to legitimize AAE, it continues to be viewed as ``bad'' English and its speakers continue to be viewed as linguistically inadequate---a view called the \emph{deficit} perspective \cite{alim2016raciolinguistics,rosa2017unsettling}. This perspective persists despite demonstrations that AAE is rule-bound and grammatical \cite{mufwene1998african,green2002african}, in addition to ample evidence of its speakers' linguistic adroitness \cite[e.g.,][]{alim2004you,rickford2016language}. This perspective belongs to a broader set of raciolinguistic ideologies \cite{rosa2017unsettling}, which also produce allocational harms; speakers of AAE are frequently penalized for not adhering to dominant language practices, including in the education system \cite{alim2004you,terry2010variable}, when seeking housing \cite{baugh2018linguistics}, and in the judicial system, where their testimony is misunderstood or, worse yet, disbelieved \cite{rickford2016language,jones2019testifying}. These raciolinguistic ideologies position racialized communities as needing linguistic intervention, such as language education programs, in which these and other harms can be reduced if communities accommodate to dominant~language practices \cite{rosa2017unsettling}.\looseness=-1

In the technology industry, speakers of AAE are often not considered consumers who matter. For example, \citet{benjamin2019race} recounts an Apple employee who worked on speech recognition for Siri:\looseness=-1

\citequotenoattr{As they worked on different English dialects --- Australian, Singaporean, and Indian English --- [the employee] asked his boss: `What about African American English?' To this his boss responded: `Well, Apple products are for the premium market.'}{} 

\noindent
The reality, of course, is that speakers of AAE tend not to represent the ``premium market'' precisely because of institutions and policies that help to maintain racial hierarchies by systematically denying them the opportunities to develop wealth that are available to white Americans \cite{rothstein2017color}---an exclusion that is reproduced in technology by countless decisions like the one described above.\looseness=-1

Engaging with the literature outlined above situates the system
behaviors that are described as ``bias,'' providing a foundation for
normative reasoning. Researchers and practitioners should be concerned
about ``racial bias'' in toxicity detection systems not only because
performance differences impair system performance, but because
they reproduce longstanding injustices of stigmatization and
disenfranchisement for speakers of AAE. In re-stigmatizing AAE, they
reproduce language ideologies in which AAE is viewed as ungrammatical,
uneducated, and offensive. These ideologies, in turn, enable
linguistic discrimination and justify enduring racial hierarchies
\cite{rosa2017unsettling}. Our perspective, which understands racial
hierarchies and raciolinguistic ideologies as structural conditions
that govern the development and deployment of technology, implies that
techniques for measuring or mitigating ``bias'' in NLP systems will
necessarily be incomplete unless they interrogate and dismantle these structural conditions, including the power relations between~technologists and racialized communities.\looseness=-1

We emphasize that engaging with the literature on AAE, racial
hierarchies in the U.S., and raciolinguistic ideologies can generate
new lines of engagement. These lines include work on the ways that the
decisions made during the development and deployment of NLP systems
produce stigmatization and disenfranchisement, and work on AAE use in
practice, such as the ways that speakers of AAE interact with NLP
systems that were not designed for them. This literature can also help
researchers and practitioners address the allocational harms that may
be produced by NLP systems, and ensure that even well-intentioned NLP
systems do not position racialized communities as needing linguistic
intervention or accommodation to dominant language practices. Finally,
researchers and practitioners wishing to design better systems can
also draw on a growing body of work on anti-racist language pedagogy
that challenges the deficit perspective of AAE and other racialized
language practices
\cite[e.g.][]{flores2018what,baker2019dismantling,martinez2019looking},
as well as the work that we described in section \ref{s:practice} on
reimagining the power relations between technologists~and~communities
affected by technology.\looseness=-1

\section{Conclusion}

By surveying 146 papers analyzing ``bias'' in NLP systems,
we found that (a) their motivations are often vague, inconsistent, and
lacking in normative reasoning; and (b) their proposed quantitative
techniques for measuring or mitigating ``bias'' are poorly matched to
their motivations and do not engage with the relevant
literature outside of NLP. To help researchers and practitioners avoid
these pitfalls, we proposed three recommendations that should guide
work analyzing ``bias'' in NLP systems, and, for each, provided several concrete research questions. These recommendations rest on a
greater recognition of the relationships between language and social
hierarchies---a step that we see~as paramount to establishing a path
forward.

\section*{Acknowledgments}

This paper is based upon work supported by the National Science Foundation Graduate Research Fellowship under Grant No. 1451512. Any opinion, findings, and conclusions or recommendations expressed in this material are those of the authors and do not necessarily reflect the views of the National Science Foundation. We thank the reviewers for their useful feedback, especially the suggestion~to include additional details about our method.

\bibliography{reading}
\bibliographystyle{acl_natbib}

\appendix

\section{Appendix} \label{s:appendix}

In Table \ref{t:quotes}, we provide examples of the papers'
motivations and techniques across several NLP tasks.\looseness=-1

\begin{table*}[p]
	\centering
		\footnotesize
		\begin{tabular}{>{\raggedright\arraybackslash}p{1.8cm}p{8.2cm}>{\raggedright\arraybackslash}p{2.1cm}>{\raggedright\arraybackslash}p{2.1cm}}
		    \toprule
		    			    & &  \multicolumn{2}{c}{\textbf{Categories}} \\
			    \cmidrule{3-4}
			\textbf{NLP task} & \textbf{Stated motivation} & \textbf{Motivations} & \textbf{Techniques} \\
			\midrule
			Language modeling \cite{bordia2019identifying} & \it ``Existing biases in data can be amplified by models and the resulting output consumed by the public \textbf{can influence them}, \textbf{encourage and reinforce harmful stereotypes}, or \textbf{distort the truth}. Automated systems that depend on these models can \textbf{take problematic actions} based on biased profiling of individuals.'' & Allocational harms, stereotyping & Questionable correlations \\ \\[-.5em]
			\rowcolor{lightgray!30!white}
			Sentiment analysis \cite{kiritchenko2018examining} & \it``Other biases can be inappropriate and result in \textbf{negative experiences} for some groups of people. Examples include, \textbf{loan eligibility and crime recidivism prediction systems}...and \textbf{resum\'{e} sorting systems} that believe that men are more qualified to be programmers than women (Bolukbasi et al., 2016). Similarly, sentiment and emotion analysis systems can also perpetuate and accentuate inappropriate human biases, e.g., systems that consider \textbf{utterances from one race or gender to be less positive} simply because of their race or gender, or customer support systems that \textbf{prioritize a call from an angry male over a call from the equally angry female}.'' & Allocational harms, other representational harms (system performance differences w.r.t. text written \emph{by} different social groups) & Questionable correlations (differences in sentiment intensity scores w.r.t. text \emph{about} different social groups) \\  \\[-.5em]
			Machine translation \cite{cho2019measuring} & \it``[MT training] may incur an association of gender-specified pronouns (in the target) and gender-neutral ones (in the source) for lexicon pairs that frequently collocate in the corpora. We claim that this kind of phenomenon seriously threatens the fairness of a translation system, in the sense that it \textbf{lacks generality} and \textbf{inserts social bias} to the inference. Moreover, the input is \textbf{not fully correct} (considering gender-neutrality) and \textbf{might offend the users} who expect fairer representations.'' & Questionable correlations, other representational harms & Questionable correlations \\  \\[-.5em]
			\rowcolor{lightgray!30!white}
			Machine translation \cite{stanovsky2019evaluating} & \it``Learned models exhibit social bias when their training data \textbf{encode stereotypes not relevant for the task, but the correlations are picked up anyway}.'' & Stereotyping, questionable correlations & Stereotyping, other representational harms (system performance differences), questionable correlations \\  \\[-.5em]
			Type-level embeddings \cite{zhao2018learning} & \it``However, embeddings trained on human-generated corpora have been demonstrated to inherit \textbf{strong gender stereotypes} that reflect social constructs....Such a bias substantially affects \textbf{downstream applications}....This concerns the practitioners who use the embedding model to build \textbf{gender-sensitive applications such as a resume filtering system or a job recommendation system} as the automated system may discriminate candidates based on their gender, as reflected by their name. Besides, biased embeddings may implicitly affect downstream applications used in our daily lives. For example, when searching for `computer scientist' using a search engine...a search algorithm using an embedding model in the backbone tends to \textbf{rank male scientists higher than females'} [sic], \textbf{hindering women from being recognized} and further exacerbating the gender inequality in the community.'' & Allocational harms, stereotyping, other representational harms & Stereotyping \\ \\[-.5em]
			\rowcolor{lightgray!30!white}
			Type-level and contextualized embeddings \cite{may2019measuring} & \it``[P]rominent word embeddings such as word2vec (Mikolov et al., 2013) and GloVe (Pennington et al., 2014) encode systematic biases against women and black people (Bolukbasi et al., 2016; Garg et al., 2018), implicating many NLP systems in scaling up social injustice.'' & Vague & Stereotyping \\  \\[-.5em]
			Dialogue generation \cite{liu2019does} & \it``Since the goal of dialogue systems is to talk with users...if the systems show discriminatory behaviors in the interactions, the user experience will be adversely affected. Moreover, public commercial chatbots can get resisted for their improper speech.'' & Vague/unstated & Stereotyping, other representational harms, questionable correlations \\
			\bottomrule
		\end{tabular}
		\caption{Examples of the categories into which the papers' motivations and proposed quantitative techniques for measuring or mitigating ``bias'' fall. Bold text in the quotes denotes the content that yields our categorizations.\looseness=-1}
		\label{t:quotes}
\end{table*}

\subsection{Categorization details}

In this section, we provide some additional details about our
method---specifically, our categorization.

\paragraph{What counts as being covered by an NLP task?} We considered a paper to cover a given NLP task if it analyzed ``bias'' with respect to that task, but not if it only evaluated overall performance on that task. For example, a paper examining the impact of mitigating ``bias'' in word embeddings on ``bias'' in sentiment analysis would be counted as covering both NLP tasks. In contrast, a paper assessing whether performance on sentiment analysis degraded after mitigating ``bias'' in word embeddings would be counted only as focusing on embeddings.

\paragraph{What counts as a motivation?} We considered a motivation to include any description of the problem that motivated the paper or proposed quantitative technique, including any normative reasoning.


We excluded from the ``Vague/unstated'' category of motivations the
papers that participated in the Gendered Ambiguous Pronoun (GAP)
Shared Task at the First ACL Workshop on Gender Bias in NLP. In an
ideal world, shared task papers would engage with ``bias'' more
critically, but given the nature of shared tasks it is understandable
that they do not. As a result, we excluded them from our counts for
techniques as well. We cite the papers here; most propose techniques
we would have categorized as ``Questionable correlations,'' with a few
as ``Other representational harms''
\cite{abzaliev2019gap,attree2019gendered,bao2019transfer,chada2019gendered,ionita2019resolving,liu2019anonymized,lois2019bert,wang2019msnet,xu2019look,yang2019fill}.

We excluded \citet{dabas2020fair} from our survey because we could not
determine what this paper's user study on fairness was actually measuring.

Finally, we actually categorized the motivation for \citet{liu2019does} (i.e., the last row in Table \ref{t:quotes}) as ``Questionable correlations'' due to a sentence elsewhere in the paper; had the paragraph we quoted been presented without more detail, we would have categorized the motivation as ``Vague/unstated.''

\subsection{Full categorization: Motivations}

\paragraph{Allocational harms}
\citet{hovy2016social,caliskan2017semantics,madnani2017building,dixon2018measuring,kiritchenko2018examining,shen2018darling,zhao2018learning,bhaskaran2019good,bordia2019identifying,brunet2019understanding,chaloner2019measuring,dearteaga2019bias,dev2019attenuating,font2019equalizing,james2019probabilistic,kurita2019measuring,mayfield2019equity,pujari2019debiasing,romanov2019name,ruane2019conversational,sedoc2019role,sun2019mitigating,zmigrod2019counterfactual,hutchinson2020social,papakyriakopoulos2020bias,ravfogel2020null,strengers2020adhering,sweeney2020reducing,tan2020its,zhang2020hurtful}.

\paragraph{Stereotyping} \citet{bolukbasi2016man,bolukbasi2016quantifying,caliskan2017semantics,mccurdy2017grammatical,rudinger2017social,zhao2017men,curry2018metoo,diaz2018addressing,santana2018there,sutton2018biased,zhao2018gender,zhao2018learning,agarwal2019word,basta2019evaluating,bhaskaran2019good,bordia2019identifying,brunet2019understanding,cao2019toward,chaloner2019measuring,cho2019measuring,dev2019attenuating,font2019equalizing,gonen2019lipstick,james2019probabilistic,kaneko2019gender,karve2019conceptor,kurita2019measuring,lauscher2019are,lee2019exploring,manzini2019black,mayfield2019equity,precenth2019word,pujari2019debiasing,ruane2019conversational,stanovsky2019evaluating,sun2019mitigating,tan2019assessing,webster2019gendered,zmigrod2019counterfactual,gyamfi2020deb2viz,hube2020debiasing,hutchinson2020social,kim2020intersectional,nadeem2020stereoset,papakyriakopoulos2020bias,ravfogel2020null,rozado2020wide,sen2020towards,shin2020neutralizing,strengers2020adhering}.

\paragraph{Other representational harms} \citet{hovy2015tagging,blodgett2016demographic,bolukbasi2016quantifying,hovy2016social,blodgett2017racial,larson2017gender,schnoebelen2017goal,blodgett2018twitter,curry2018metoo,diaz2018addressing,dixon2018measuring,kiritchenko2018examining,park2018reducing,shen2018darling,thelwall2018gender,zhao2018learning,badjatiya2019stereotypical,bagdasaryan2019differential,bamman2019annotated,cao2019toward,chaloner2019measuring,cho2019measuring,davidson2019racial,dearteaga2019bias,fisher2019measuring,font2019equalizing,garimella2019gender,loukina2019many,mayfield2019equity,mehrabi2019man,nozza2019unintended,prabhakaran2019perturbation,romanov2019name,ruane2019conversational,sap2019risk,sheng2019woman,sun2019mitigating,sweeney2019transparent,vaidya2019empirical,gaut2020towards,gencoglu2020cyberbullying,hovy2020can,hutchinson2020social,kim2020intersectional,peng2020fine-tuning,rios2020fuzze,sap2020social,shah2020predictive,sheng2020towards,tan2020its,zhang2020demographics,zhang2020hurtful}.

\paragraph{Questionable correlations} \citet{jorgensen2015challenges,hovy2016social,madnani2017building,rudinger2017social,zhao2017men,burns2018women,dixon2018measuring,kiritchenko2018examining,lu2018gender,park2018reducing,shen2018darling,zhang2018mitigating,badjatiya2019stereotypical,bhargava2019exposing,cao2019toward,cho2019measuring,davidson2019racial,dev2019measuring,garimella2019gender,garg2019counterfactual,huang2019reducing,james2019probabilistic,kaneko2019gender,liu2019does,karve2019conceptor,nozza2019unintended,prabhakaran2019perturbation,romanov2019name,sap2019risk,sedoc2019role,stanovsky2019evaluating,sweeney2019transparent,vaidya2019empirical,zhiltsova2019mitigation,chopra2020hindi,gonen2020automatically,gyamfi2020deb2viz,hube2020debiasing,ravfogel2020null,rios2020fuzze,ross2020measuring,saunders2020reducing,sen2020towards,shah2020predictive,sweeney2020reducing,yang2020causal,zhang2020demographics}.


\paragraph{Vague/unstated} \citet{rudinger2018gender,webster2018mind,dinan2019queens,florez2019unintended,jumelet2019analysing,lauscher2019general,liang2019towards,maudslay2019name,may2019measuring,prates2019assessing,prost2019debiasing,qian2019reducing,swinger2019what,zhao2019gender,zhou2019examining,ethayarajh2020is,huang2020multilingual,jia2020mitigating,popovic2020joint,pryzant2020automatically,vig2020causal,wang2020double,zhao2020gender}.

\paragraph{Surveys, frameworks, and meta-analyses} \citet{hovy2016social,larson2017gender,mccurdy2017grammatical,schnoebelen2017goal,basta2019evaluating,ethayarajh2019understanding,gonen2019lipstick,lauscher2019are,loukina2019many,mayfield2019equity,mirzaev2019considerations,prabhumoye2019principled,ruane2019conversational,sedoc2019role,sun2019mitigating,nissim2020fair,rozado2020wide,shah2020predictive,strengers2020adhering,wright2020recast}.

\section{Full categorization: Techniques}

\paragraph{Allocational harms}
\citet{dearteaga2019bias,prost2019debiasing,romanov2019name,zhao2020gender}.

\paragraph{Stereotyping} \citet{bolukbasi2016man,bolukbasi2016quantifying,caliskan2017semantics,mccurdy2017grammatical,diaz2018addressing,santana2018there,sutton2018biased,zhang2018mitigating,zhao2018gender,zhao2018learning,agarwal2019word,basta2019evaluating,bhaskaran2019good,brunet2019understanding,cao2019toward,chaloner2019measuring,dev2019attenuating,ethayarajh2019understanding,gonen2019lipstick,james2019probabilistic,jumelet2019analysing,kaneko2019gender,karve2019conceptor,kurita2019measuring,lauscher2019are,lauscher2019general,lee2019exploring,liang2019towards,liu2019does,manzini2019black,maudslay2019name,may2019measuring,mirzaev2019considerations,prates2019assessing,precenth2019word,prost2019debiasing,pujari2019debiasing,qian2019reducing,sedoc2019role,stanovsky2019evaluating,tan2019assessing,zhao2019gender,zhou2019examining,chopra2020hindi,gyamfi2020deb2viz,nadeem2020stereoset,nissim2020fair,papakyriakopoulos2020bias,popovic2020joint,ravfogel2020null,ross2020measuring,rozado2020wide,saunders2020reducing,shin2020neutralizing,vig2020causal,wang2020double,yang2020causal,zhao2020gender}.

\paragraph{Other representational harms} \citet{jorgensen2015challenges,hovy2015tagging,blodgett2016demographic,blodgett2017racial,blodgett2018twitter,curry2018metoo,dixon2018measuring,park2018reducing,thelwall2018gender,webster2018mind,badjatiya2019stereotypical,bagdasaryan2019differential,bamman2019annotated,bhargava2019exposing,cao2019toward,font2019equalizing,garg2019counterfactual,garimella2019gender,liu2019does,loukina2019many,mehrabi2019man,nozza2019unintended,sap2019risk,sheng2019woman,stanovsky2019evaluating,vaidya2019empirical,webster2019gendered,ethayarajh2020is,gaut2020towards,gencoglu2020cyberbullying,hovy2020can,huang2020multilingual,kim2020intersectional,peng2020fine-tuning,ravfogel2020null,rios2020fuzze,sap2020social,saunders2020reducing,sheng2020towards,sweeney2020reducing,tan2020its,zhang2020demographics,zhang2020hurtful}.

\paragraph{Questionable correlations} \citet{jurgens2017incorporating,madnani2017building,rudinger2017social,zhao2017men,burns2018women,diaz2018addressing,kiritchenko2018examining,lu2018gender,rudinger2018gender,shen2018darling,bordia2019identifying,cao2019toward,cho2019measuring,davidson2019racial,dev2019measuring,dinan2019queens,fisher2019measuring,florez2019unintended,font2019equalizing,garg2019counterfactual,huang2019reducing,liu2019does,nozza2019unintended,prabhakaran2019perturbation,qian2019reducing,sap2019risk,stanovsky2019evaluating,sweeney2019transparent,swinger2019what,zhiltsova2019mitigation,zmigrod2019counterfactual,hube2020debiasing,hutchinson2020social,jia2020mitigating,papakyriakopoulos2020bias,popovic2020joint,pryzant2020automatically,saunders2020reducing,sen2020towards,shah2020predictive,sweeney2020reducing,zhang2020hurtful}.


\paragraph{Vague/unstated} None.

\paragraph{Surveys, frameworks, and meta-analyses} \citet{hovy2016social,larson2017gender,mccurdy2017grammatical,schnoebelen2017goal,basta2019evaluating,ethayarajh2019understanding,gonen2019lipstick,lauscher2019are,loukina2019many,mayfield2019equity,mirzaev2019considerations,prabhumoye2019principled,ruane2019conversational,sedoc2019role,sun2019mitigating,nissim2020fair,rozado2020wide,shah2020predictive,strengers2020adhering,wright2020recast}.

\end{document}